\pdfoutput=1

\documentclass[11pt]{article}

\usepackage[]{acl}

\usepackage{times}
\usepackage{latexsym}

\usepackage[T1]{fontenc}

\usepackage[utf8]{inputenc}

\usepackage{microtype, graphicx}
\usepackage{float}
\usepackage{booktabs}
\title{Divide and Prompt: Chain of Thought Prompting for Text-to-SQL}

\author{ Xiping Liu, Zhao Tan\\
  Jiangxi University of Finance and Economics}

\begin{document}
\maketitle
\begin{abstract}
Chain-of-thought (CoT) prompting combined with large language models (LLMs) have achieved encouraging results on complex reasoning tasks. Text-to-SQL is a critical semantic parsing task that converts natural language questions into SQL statements, involving a complex reasoning process. However, there is little work about using CoT prompting to activate LLM's reasoning capabilities on Text-to-SQL tasks. In this work, we propose a new paradigm for prompting Text-to-SQL tasks, called Divide-and-Prompt, which first divides the task into subtasks, and then approach each subtask through CoT. We present 3 prompting-based methods to enhance the Text-to-SQL ability of LLMs. Experiments show that these prompts guide LLMs to generate Text-to-SQL with higher execution accuracy.
\end{abstract}

\section{Introduction}

With the increasing size of large language models (LLMs), they excel at various natural language processing tasks and have become an essential element of natural language processing. Models such as BERT\cite{devlin2019bert}, BART\cite{lewis2019bart}, and T5\cite{raffel2020exploring} require fine-tuning with a small amount of relevant data. However, their fine-tuning becomes very costly as the models' size grows. Models such as GPT-3\cite{brown2020language}, LaMDA\cite{thoppilan2022lamda}, and PaLM\cite{chowdhery2022palm} require prompt design to generate target outputs. ChatGpt\footnote
{\url{https://chat.openai.com/}}has a powerful performance in (in-context) few-shot and zero-shot learning by employing Reinforcement Learning for Human Feedback (RLHF)\cite{christiano2023deep}. However, scaling up the model size has yet to prove sufficient for achieving high performance on challenging tasks such as Text-to-SQL\cite{liu2023comprehensive,rajkumar2022evaluating}.

\begin{figure}
     \centering
     \includegraphics[width=0.45\textwidth]{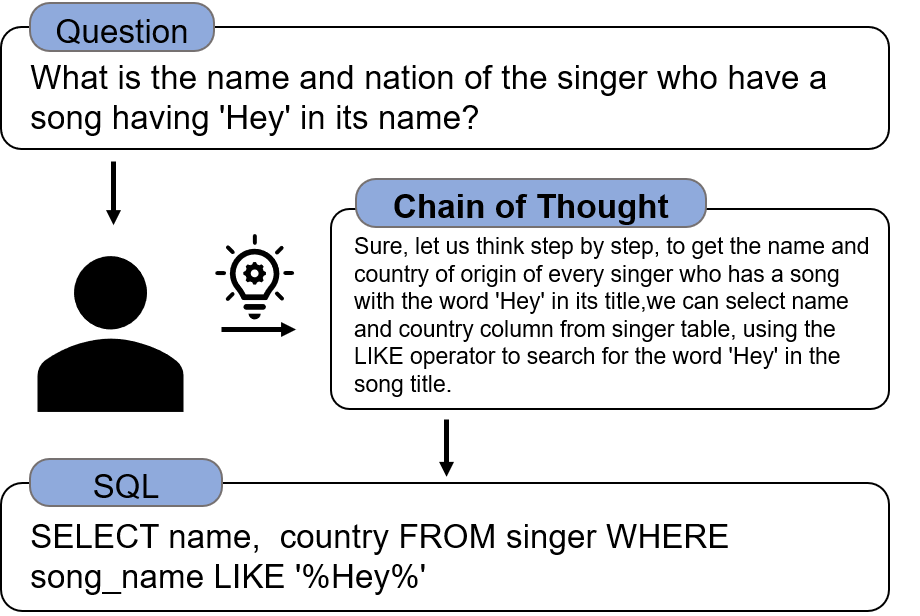}
     \caption{This is an illustration of the inference process in Text-to-SQL.}
     \label{fig:example for text to sql}
 \end{figure}

Translating text to SQL is a challenging process involving extensive and complex reasoning. As show in Figure \ref{fig:example for text to sql}. 

It is promising to consider Text-to-SQL as a reasoning task. We notice that CoT prompting\cite{wei2023chainofthought,kojima2023large} induces LLMs to produce a series of intermediate steps before the final answer to a question; CoT prompting elicits challenging tasks such as arithmetic, commonsense, and symbolic reasoning in LLMs. However, as shown in table \ref{tab:main experiment}, normal CoT prompting does not perform well on Text-to-SQL. This is probably because the task involves much reasoning steps, concerning the understanding of the query intentions as well as the database schema.

Inspired by the characteristics of Text-to-SQL, we propose a new paradigm for prompting Text-to-SQL tasks, called Divide-and-Prompt (DnP). The basic idea is to divide the task into subtasks, and then tackle each subtask through CoT. We design 3 DnP promptings for Text-to-SQL and evaluate them on LLMs. Based on the experimental results, the following findings were obtained:
 \begin{itemize}
    \item DnP promptings is very effective for Text-to-SQL. Compared with the standard zero-shot prompt, our proposed prompts improve execution accuracy by \textbf{4.3\%} .
    \item DnP promptings is specially useful for difficult Text-to-SQL. On hard-level Text-to-SQL tasks, our prompts improved the execution accuracy by up to \textbf{10.8\%}; for extra-level tasks, it also showed a 3\% improvement. 
    \item Normal CoT prompting does not perform well on SQL generation. Due to the strict structure and syntax of SQL, normal CoT prompting hard to inducing the helpful reasoning chains for Text-to-SQL in LLMs.
\end{itemize}

\section{Method}
\begin{figure*}
    \centering
    \includegraphics[width=\linewidth,scale=1.00]{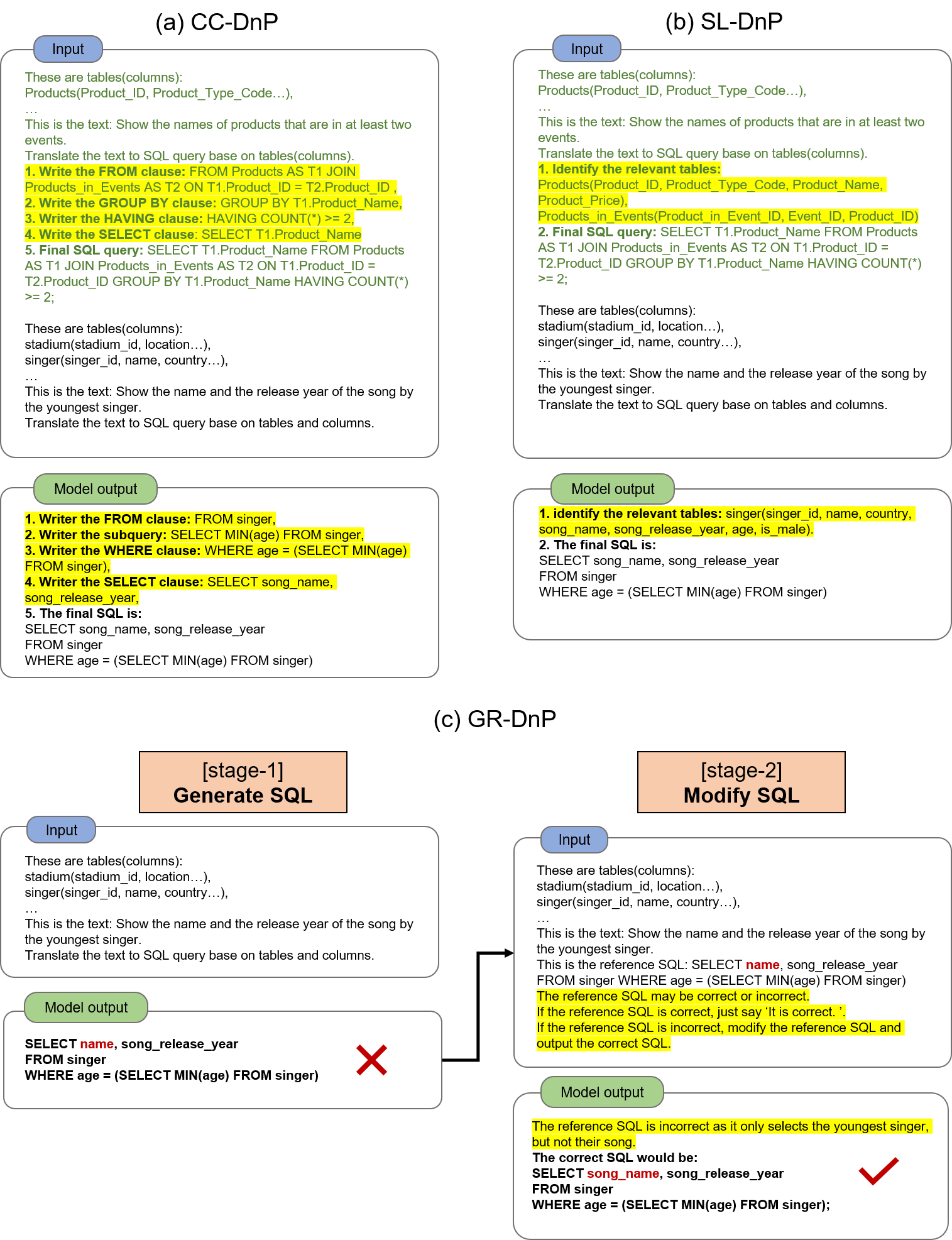}
    \caption{This is an illustration of the methods: (a) Clause by Clause DnP (CC-DnP), (b) Schema Linking DnP (SL-DnP), (c) Generate and Refine DnP(GR-DnP). The demonstrations in few-shot are green text, and the reasoning steps in model input and output are highlighted.}
    \label{fig:Example of methods}
\end{figure*}

In this work, we propose a new paradigm for prompts of Text-to-SQL, called Divide-and-prompt (DnP). The basic idea is to instruct the model to divide complex tasks into subtasks, and then solve each subtasks. There are different ways of dividing a Text-to-SQL task, therefore, there are many possible DnP methods. We designed 3 DnP methods, as shown in Figure \ref{fig:Example of methods}, which show effectiveness on Text-to-SQL task. Note that we used only natural language to construct the prompts. This is because  Text-to-SQL technology is expected to be used by non-expert users; making prompts easier to understand for users is essential.

~\\\textbf{Clause by Clause DnP (CC-DnP).} In this method, the model is induced to generates an SQL query clause by clause, e.g. first generate the SELECT clause, then FROM clause, $\cdots$, as shown in Figure \ref{fig:ablation}. We found that the order in which clauses are generated is an essential factor that impacts the results, and we discuss this in \ref{sec:ablation}.

~\\\textbf{Schema Linking DnP (SL-DnP).} \cite{li2023resdsql} has confirmed that schema linking, i.e. identifying relevant schema elements (tables, columns, etc.), is valuable for the model to generate SQL correctly. We proposed SL-DnP method, by which the model first learn to identify the relevant schema elements relevant to the question, and then generate the SQL. What's more, we found that different ways of schema linking have impacts on the performance. More discussions can be found in \ref{sec:ablation}.

~\\\textbf{Generate and Refine DnP(GR-DnP).} Recently, there have been some attempts\cite{madaan2023selfrefine}  to use LLMs to modify the raw output that may have some mistakes. In this work, we propose the GR-DnP method, which generate an SQL query in two stages. In the first stage, the model generates an initial SQL; In the second stage, the model checks and refines the SQL as needed. Figure \ref{fig:Example of methods} shows an example.

\section{Experiment}

\subsection{Experiment Setup}

~\\\textbf{Model.}  Our experiment focuses on the models accessible via the OpenAI API: GPT-3.5-Turbo. We also conducted experiments on other GPT-3.5 models(text-davinci-002, text-davinci-003, i.e.). However, their Text-to-SQL capability is far inferior to GPT-3.5-Turbo. The model parameters we set for \textit{temperature} and \textit{top\_p} are 0.3 and 1, respectively.

~\\\textbf{Datasets.}	We conduct experiments on the standard Spider dataset\cite{yu2019spider}, a large-scale cross-domain Text-to-SQL benchmark containing 8659 training samples across 146 databases and 1034 evaluation samples across 20 databases. The data richness of the Spider dataset is sufficient to verify the validity of the methods designed in this paper.


~\\\textbf{Evaluation Metrics.} The most popular evaluation metrics for Text-to-SQL are Exact Match(EM) and Execution Accuracy(EX). EM evaluates whether the generated queries exactly match the golden answers, while EX evaluates the correctness of generated answers based on the execution results.
In this work, we do not adopt EM, because we think that SQL is flexible in syntax, thus EX makes much sense. What is more, GPT-3.5 was not fine-tuned on the Spider dataset, it tends to generate SQL queries with a lot of variety different from the golden answers. We also adopt valid SQL(VA) and test-suite accuracy (TS)\cite{zhong2020semantic} as evaluation metrics. All results are the average of three experiment results.

~\\\textbf{Baselines.} We primarily utilized the following baselines: (1) PICARD \cite{scholak2021picard} is a method for constraining auto-regressive decoders of language models through incremental parsing. (2) Graphix-T5 \cite{li2023graphixt5} propose a mixed model with the standard pre-trained transformer model augmented by some specially-designed graph-aware layers. (3) RESDSQL \cite{li2023resdsql} proposes a ranking-enhanced encoding and skeleton-aware decoding framework to decouple the schema linking and the skeleton parsing. (4) \cite{rajkumar2022evaluating} perform an empirical evaluation of Text-to-SQL capabilities of the Codex language model. (5) \cite{liu2023comprehensive} presents the comprehensive analysis of ChatGPT’s Text-to-SQL ability. 

\subsection{Main Experiment}

\textbf{Overall Performance}  We use natural language to construct prompts. In the zero-shot scenario, the EX performance of GPT-3.5 is 70.8\%. We construct normal CoT, CC-DnP, SL-DnP, and GR-DnP promoting in the few-shot learning scenario, and the demonstrations in the few-shot are the clustering results of question in the Spider training set.

We construct demonstrations for CoT promopting by requiring the model to \textit{think step by step}. Manually construct demonstrations for CC-DnP, SL-DnP, and GR-DnP promopting.

The experiment results show that our methods improve execution accuracy by 4.3\% . Despite there being a gap (9\%) in EX compared to the current SOTA model\cite{li2023resdsql}, it is remarkable that GPT-3.5 achieved 75.1\% EX with GR-DnP prompting considering that it was not finetuned on the Spider.

\begin{table}[H]
\centering
\begin{tabular}{llll}
\toprule[2pt]
\textbf{Methods} & \textbf{VA} & \textbf{EX} & \textbf{TS} \\
\midrule[1pt]
\textit{Finetuned} \\
T5-3B + PICARD & 98.4 & 79.3 & 69.4\\
GRAPHIX + PICARD & 98.8 & 80.5 & 70.3 \\
RESDSQL + NatSQL & 99.1 & \textbf{84.1} & 73.5 \\ 
\midrule[1pt]
\textit{Prompting only} \\
\cite{rajkumar2022evaluating} & 91.6 & 67.0 & 55.1 \\ 
\cite{liu2023comprehensive} & 97.7 & \textbf{70.1} & 60.1 \\
\midrule[1pt]
\textit{ours} & & &\\ 
GPT-3.5 (zero-shot)  & 97.9  & 70.8 & 62.3 \\
GPT-3.5 (few-shot) & 98.2 & 72.9 & 62.6 \\
GPT-3.5 + normal CoT  & 92.6  & 60.3 & 49.5 \\
GPT-3.5 + CC-DnP  & 97.8  & 74.3 & 63.0 \\
GPT-3.5 + RL-DnP  & 99.1  & 74.7 & 65.1 \\
GPT-3.5 + GR-DnP  & 98.6  & \textbf{75.1} & 65.4 \\
\bottomrule[2pt]
\end{tabular}
\caption{Prior best Spider development set performance across models, as measured by percentage of predictions which are valid SQL (VA), execution accuracy (EX), test-suite accuracy (TS).}
\label{tab:main experiment}
\end{table}

~\\\textbf{Results on Complex Queries.}  We compare the more precise performance results of our prompts in four separate SQL difficulty levels separated by
Spider officially. As \cite{wei2023chainofthought} pointed out, CoT promptings are particularly helpful for complex reasoning tasks.

It is evident in Table \ref{tab:tab2}. For difficult Text-to-SQL tasks, our prompts stimulate the potential of the model.
For hard-level Text-to-SQL tasks, GR-DnP prompting improved the EX by 10.8\% compared to standard prompting, which is encouraging.

\begin{table*}
\centering
\begin{tabular}{lccccc}
\hline
\textbf{Prompts} & \textbf{Easy} & \textbf{Medium} & \textbf{Hard} & \textbf{Extra} & \textbf{all} \\
\hline
 Standard Prompting & 91.1  & 78.5 & 58.0 & 46.4  & 72.9 \\
 Normal CoT Prompting  & 71.4 & 62.8 & 52.9 & 45.2 & 60.3 \\
 CC-DnP  & 89.1 & 79.1 & 65.5 & 48.2 & 74.3 \\ 
 LR-DnP  & \textbf{91.5} & 78.8 & 64.2 & 49.4 & 74.7 \\ 
 GR-DnP  & 89.9 & \textbf{79.1} & \textbf{68.8(10.8 $\uparrow$ )} & \textbf{49.4} & \textbf{75.1} \\ 
\hline
\end{tabular}
\caption{Execution accuracy (EX) by varying the levels of difficulty of the inference data.}
\label{tab:tab2}
\end{table*}

\begin{figure}  
    \centering
    \includegraphics[width=\linewidth,scale=1.00]{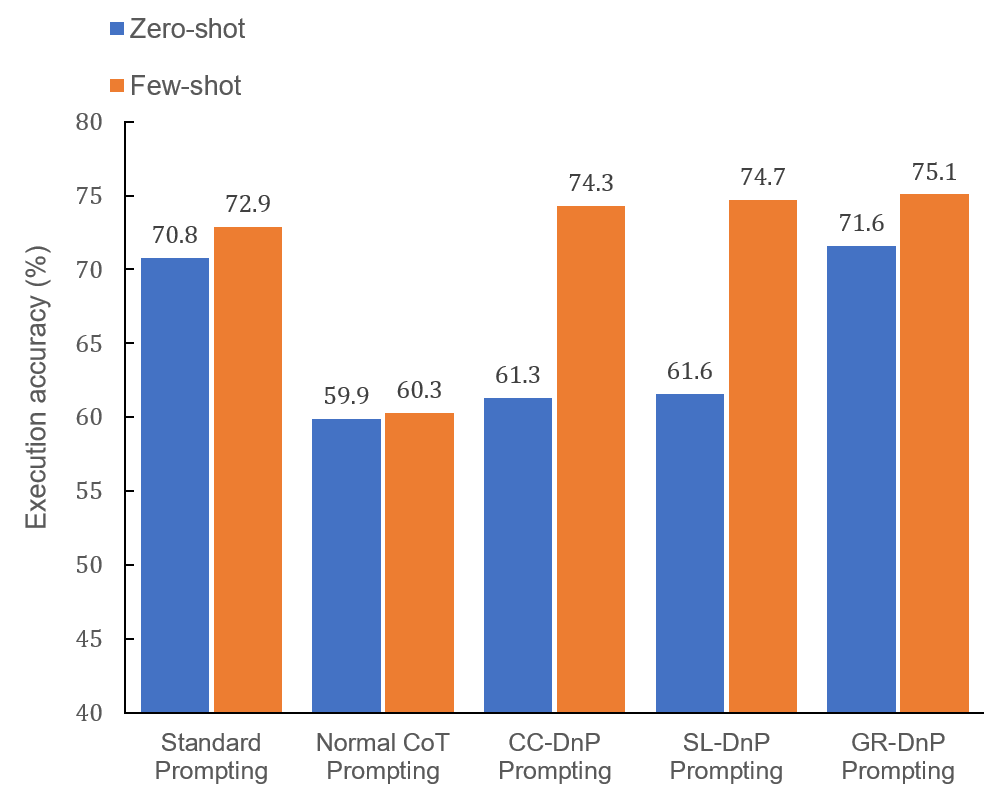}
    \caption{Performance of zero-shot learning and few-shot learning of prompts.}
    \label{fig:Performance of zero-shot learning and few-shot learning of prompts}
\end{figure}

~\\\textbf{Results of zero-shot learning.}  We compared the performance of prompts in zero-shot and few-shot learning scenarios. We have designed corresponding zero-shot prompts for our methods; the results are shown in Figure \ref{fig:Performance of zero-shot learning and few-shot learning of prompts}. 

For Text to SQL, designing a effective zero-shot prompt is very challenging. In the zero-shot scenario, the performance of all prompts was greatly reduced; in LR-DnP, the performance declined by 13.1\%.

By observing the model's output, in the zero-shot scenario, the model does not follow the reasoning steps provided by the prompt. CC-DnP, LR-DnP, and normal CoT prompting have no significant difference for performers in the zero-shot scenario.

\subsection{Ablation Study}
\label{sec:ablation}

\begin{figure*}
     \centering
     \includegraphics[width=0.9\textwidth]{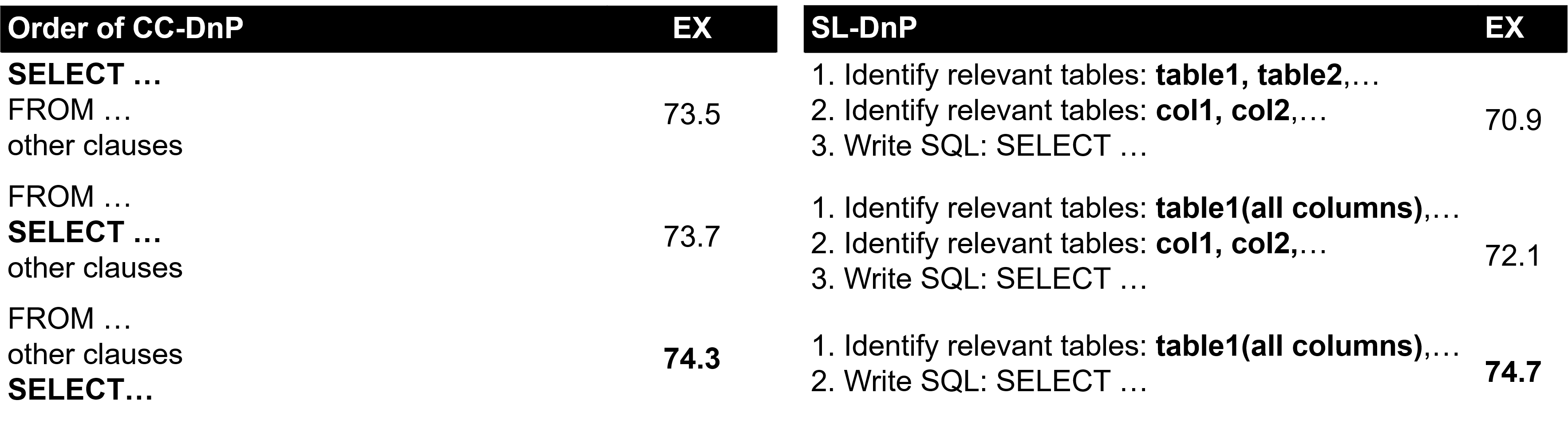}
     \caption{Execution accuracy (EX) of different CC-DnP and LR-DnP.}
     \label{fig:ablation}
 \end{figure*}

 \textbf{Ablation study of CC-DnP.}	
The order of clause generation is essential, and we have experimented with three different orders of prompts, as shown in Figure \ref{fig:ablation}.

Different orders represent different reasoning paths. For example, the SELECT clause first represents thinking in SQL generating standard order. However, this is counterintuitive because when writing an SQL, we must first consider which tables the SQL involves rather than which columns.

Due to SQL queries can be expressed in various ways, it is challenging to define an optimal order. This paper considers 3 orders and concludes that the SELECT clause last is the most suitable order.

~\\\textbf{Ablation study of SL-DnP.}  	
Finding relevant schema elements before generating SQL is an intuitive idea that is almost impossible. We want to guide the model via SL-DnP prompting to identify relevant tables or columns, which ultimately helps generate the target SQL.

For SL-DnP prompting, it is necessary to consider whether it is necessary to identify each related table and column. Because identifying tables and columns are inherently challenging for the LLMs. Based on this, we propose 3 SL-DnP are shown in Figure \ref{fig:ablation}.

The experiment results show that identifying which schema elements significantly impact the results. The first prompt in Figure \ref{fig:ablation} induces the model to identify each relevant table and column name, resulting in the worst performance. However, less precise prompts have better performance, such as the third prompt in the Figure \ref{fig:ablation} only guiding the model to find the relevant table and all the columns in this table, achieving the best performance.

\begin{figure} 
    \centering
    \includegraphics[width=\linewidth,scale=1.00]{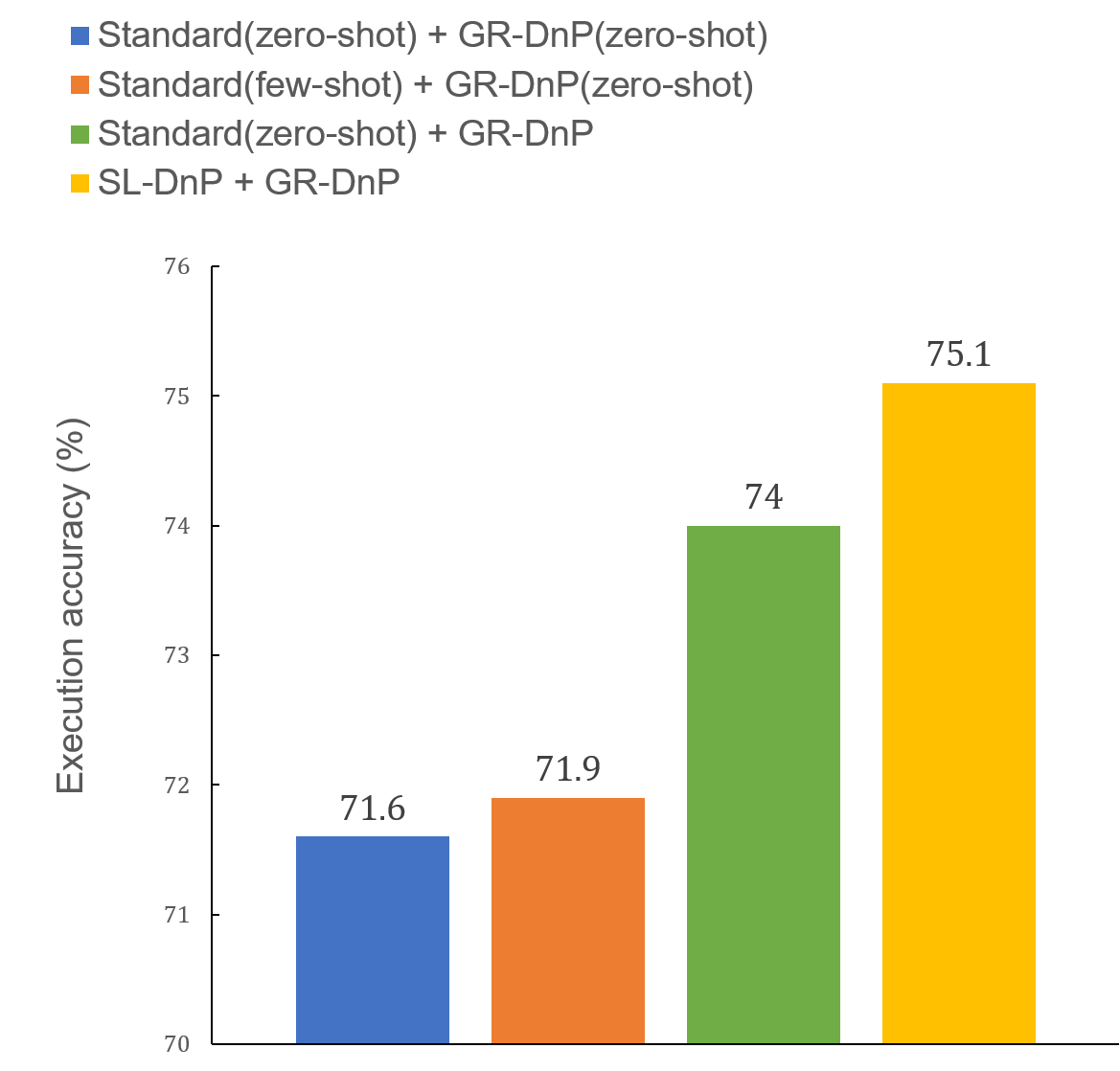}
    \caption{Ablation study of GR-DnP. SL-DnP + GR-DnP means generating SQL with SL-DnP in stage-1 and refining SQL with GR-DnP in stage-2.}
    \label{fig:TMCoT ablation}
\end{figure}

~\\\textbf{Ablation study of GR-DnP.}		
For the GR-DnP, model generate SQL in stage-1, and in stage-2, checke and modify SQL as needed. We compared the performance of two stages in zero-shot and few-shot learning scenarios, respectively, as shown in figure\ref{fig:TMCoT ablation}.

\section{Related Work}
This section reviews two lines of research that form the basis of this work: CoT prompting and Text-to-SQL task.

\subsection{Chain of Thought Prompting}
These are two primary paradigms for CoT prompting. One is called zero-shot-CoT\cite{kojima2023large}, adding a single prompt like \textit{“Let’s think step by step” } before the answer to inducing the reasoning chains in LLMs. The other paradigm is few-shot prompting with manual or model-generated reasoning demonstrations\cite{wei2023chainofthought}. Each demonstration has a question and a reasoning chain. A reasoning chain comprises a  series of intermediate reasoning steps and an expected answer. CoT prompting enhances perform of LLMs in challenging reasoning tasks.

\subsection{Text-to-SQL}
Text-to-SQL is a task that translates natural language questions posed by non-expert users into SQL statements. Spider dataset\cite{yu2019spider} collects many natural language questions and their corresponding SQL and covers many complex language structures and operations. Based on Spider, many other challenging datasets have been proposed(Spider-SYN\cite{gan-etal-2021-towards}, Spider-DK\cite{gan2021exploring}, Spider-CG\cite{gan2022measuring}).

From a modeling viewpoint, two distinct approaches are often utilized in Text-to-SQL. One is employing Graph Neural Networks to utilize the structural information of text and schema(e.g., RATSQL\cite{wang2021ratsql}, Graphix-T5\cite{li2023graphixt5}). Another is the use of pre-trained models (e.g., T5\cite{raffel2020exploring}, GAP\cite{shi2020learning}); LLMs have been applied in Text-to-SQL, and PICARD\cite{scholak2021picard} utilizes the T5-3B model but still requires the training data for fine-tuning. \cite{rajkumar2022evaluating}investigated the Text-to-SQL capabilities of the GPT3 model. \cite{liu2023comprehensive}evaluate the comprehensive Text-to-SQL capabilities of ChatGPT. 

However, prior works employed the direct prompt, which only partially exploits the LLMs' capabilities. To the best of our knowledge, there is currently no work to explore CoT prompting of Text-to-SQL, and we are the first to connect Text-to-SQL with reasoning task employing CoT prompting to enhance LLMs' ability to generate Text-to-SQL.

\section{Discussion}

\textbf{Why few-shot learning? }
Experiment results show that our prompts have impressive performance only in the few-shot learning scenario. We observe the model's output in the zero-shot scenario, and observe that the model does not strictly follow the reasoning steps required in the prompts. Zero-shot prompting cannot guide the model to reason as required.

~\\\textbf{Demonstrations in few-shot learning.}
The demonstrations in the few-shot have a significant effect on the results\cite{wei2023larger}; they should be the representative. We cluster the question in Spider training set into 5 clusters\cite{zhang2022automatic}. The clustering result is 5 questions that start with “what, find, how, which, and show.” However, we expect the demonstrations to represent various query methods instead of questioning methods. We will find more suitable methods to generate demonstrations for Text-to-SQL in the future.

\section{Conclusion}

This paper considers Text-to-SQL as a reasoning task and proposes 3 prompting-based methods to enhance LLMs' ability to generate Text-to-SQL. SQL statements have strict syntax and structure, and normal CoT prompting cannot induce LLMs to generate Text-to-SQL well. We have designed CC-DnP, SL-DnP, and GR-DnP prompting for Text-to-SQL based on Text-to-SQL characteristics to induce LLMs to make helpful reasoning chains. Compared to the standard prompting, our proposed prompts improve execution accuracy by 4.3\%, especially for hard-level Text-to-SQL tasks, improving execution accuracy by 10.8\%.

\bibliography{anthology,custom}
\bibliographystyle{acl_natbib}




\end{document}